\documentclass[preprint,12pt]{elsarticle}
\makeatletter
\def\ps@pprintTitle{%
   \let\@oddhead\@empty
   \let\@evenhead\@empty
   \let\@oddfoot\@empty
   \let\@evenfoot\@empty
}
\makeatother

\usepackage{amsmath,amssymb}
\usepackage{booktabs}
\usepackage{tikz}
\usepackage{pgfplots}
\pgfplotsset{compat=1.18}
\usepackage{graphicx}
\usepackage{algorithm}
\usepackage{algorithmic}
\usepackage{adjustbox}
\journal{}

\begin{document}
\begin{frontmatter}

\title{FlowOE: Imitation Learning with Flow Matching for Optimal Execution under Heston Volatility and Concave Market Impacts}

\author[inst1]{Yang Li}\corref{cor1}
\ead{yli269@stevens.edu}
\author[inst1]{Zhi Chen}

\address[inst1]{Department of Financial Engineering \\
Stevens Institute of Technology \\
Hoboken, NJ, USA \\
\{yli269,zchen100\}@stevens.edu
}
\cortext[cor1]{Corresponding author}

\begin{abstract}
Optimal execution in financial markets refers to the process of strategically transacting a large volume of assets over a period to achieve the best possible outcome by balancing the trade-off between market impact costs and timing or volatility risks.
Traditional optimal execution strategies, such as static Almgren-Chriss models, often prove suboptimal in dynamic financial markets. This paper proposes FlowOE, a novel imitation learning framework based on flow matching models, to address these challenges. FlowOE learns from a diverse set of expert traditional strategies and adaptively selects the most suitable expert behavior for prevailing market conditions. To the best of our knowledge, this work is the first to apply flow matching models in a stochastic optimal execution problem. Empirical evaluations across various market conditions demonstrate that FlowOE significantly outperforms both the specifically calibrated expert models and other traditional benchmarks, achieving higher profits with reduced risk. These results underscore the practical applicability and potential of FlowOE to enhance adaptive optimal execution.
\end{abstract}


\end{frontmatter}

\section{Introduction}
Optimal execution (OE) in financial markets is a strategy for transacting large volumes of assets over a period to achieve the best possible outcome. Concurrently, OE aims to minimize execution risk, with a particular focus on adverse price movements, known as market impact \cite{obizhaeva2013optimal, donnelly2022optimal, almgren2001optimal}. 
Optimal execution is typically driven by needs such as implementing investment strategies, rebalancing portfolios, responding to new market information or opportunities, managing client cash flows and mandates, or simply the operational scale of their extensive assets.
Therefore, OE is crucial for both investors achieve better returns at lower costs, and the market benefits from enhanced liquidity, price discovery, and stability.

Traditional approaches such as Almgren-Chriss model \cite{almgren2001optimal} formulate the OE task as a stochastic control problem and solve it with analytical form. However, these methods take simplifying assumptions \cite{gatheral2011optimal}. For example, it considers market volatility as a fixed constant throughout the execution period, which contrasts with real-world scenarios where volatility is dynamic and can change unpredictably. Furthermore, it commonly assumes that market impact scales linearly with the size or speed of trades. In actual markets, this relationship is often non-linear \cite{ma2020class,almgren2003optimal}. Typically, as the execution of a large order progresses and cumulatively consumes liquidity, the marginal impact of subsequent trades can increase nonlinearly. Given these challenges in traditional models, this paper aims to explore a framework to address the optimal execution problem under more practical and dynamic market conditions. We aim to find an adaptive optimal execution strategies in complex markets. 

Traditional model, which pre-calibrate the parameters, often prove suboptimal in dynamic market conditions because they cannot adjust to real-time changes in factors like liquidity and volatility. To address this rigidity and create strategies that can better adapt to the market's evolving state, researchers have increasingly turned to reinforcement learning (RL) \cite{macri2025reinforcement}. RL allows agents to learn optimal trading policies by interacting with the market environment, making them capable of navigating complex and time-varying liquidity. Several studies have demonstrated that RL-based approaches can outperform traditional models by adapting execution based on prevailing market conditions \cite{lorenz2011mean}. 
While Reinforcement Learning agents can often be trained to achieve high performance and learn sophisticated strategies within a specific training environment, such as a market simulator or on a particular historical dataset, a critical challenge arises when these agents are deployed in new or live financial market conditions. The inherent stochasticity and high levels of noise characteristic of financial markets mean that patterns and policies learned during training may not generalize effectively. Consequently, these RL agents can exhibit a lack of robustness, with their performance degrading significantly when faced with real-world market dynamics, subtle regime shifts, or out-of-distribution events not encountered in their original training setting.

Flow matching models, recent generative modeling advancements \cite{lipman2022flow, tong2023conditional, liu2022rectified}, excel at learning complex data distributions and transformations, a capability leveraged in robotic imitation learning. In robotics, this allows machines to acquire sophisticated skills by observing human expert demonstrations, making robot behavior more predictable and safer, especially for complex tasks.
This paradigm compellingly transfers to finance, particularly for optimal execution in stochastic environments. Here, ``experts'' are not humans but well-established financial models (e.g., Almgren-Chriss, RL agents) or strategies optimized for specific market conditions. The flow matching model learns to map market states to optimal action distributions by observing these diverse financial experts. This transfer is reasonable because, akin to robots benefiting from human intuition, a financial trading model can learn from the distilled "wisdom" of specialized expert models to navigate complex and dynamic markets. The generative nature of flow matching is particularly adept at capturing the multi-modal distributions of optimal actions often required in such stochastic financial settings.
\cite{li2025flowhft} first explore the application of an imitation learning framework to high-frequency trading tasks and demonstrates that flow matching-based models are capable of learning effectively within stochastic financial environments.

To address these aforementioned challenges and motivated by insights from relevant prior studies, we propose an imitation learning framework FlowOE, which leverages a flow matching model. This framework learns from a set of expert traditional strategies, such as Almgren-Chriss models specifically calibrated to deliver optimal performance under diverse market scenarios. The core idea is that FlowOE, as a single unified model, can learn to identify the current market conditions and adaptively select or emulate the most suitable expert strategy. 
Furthermore, our ultimate aim is to develop a system that can achieve performance superior to that of the expert. We introduce a fine-tuning process that searches an adjustment action. This adjustment is then added to the action initially generated by the pre-trained imitation learning model. This additive methodology allows the system to use the expert's strategy as a strong baseline while iteratively obtaining improvements, thereby enabling the model to explore refinements and achieve a higher level of performance against the underlying task objectives than the expert alone.

To evaluate the practical applicability of our proposed FlowOE framework, we conducted comprehensive empirical analyses across diverse market conditions. We benchmarked FlowOE's performance against not only the specific expert traditional models (e.g., Almgren-Chriss) calibrated for each respective environment but also against other standard traditional strategies. The results from these experiments are compelling: FlowOE consistently and significantly outperformed the expert models, even those specifically tuned for the given conditions. This indicates that our framework successfully learns from the expert strategies and, more importantly, refines their actions effectively. Across all tested scenarios, FlowOE surpassed each benchmark model, demonstrating an ability to secure higher profits while undertaking less risk. These findings strongly illustrate our model's capability for practical deployment in real-world trading. To the best of our knowledge, this work represents the first application of a flow matching model to address challenges within a financial stochastic environment.

\section{Literature Review}
Optimal execution exists as a direct consequence of market frictions. In an idealized, perfectly liquid, and frictionless market, any order, regardless of size, could be executed instantaneously at the prevailing market price \cite{donnelly2022optimal}. However, real-world markets are characterized by ``liquidity rationing'', meaning that liquidity is not infinite and large orders face constraints. Furthermore, traders encounter both direct trading costs (e.g., commissions) and indirect trading costs such as the bid-ask spread and the price impact of their own trades \cite{skachkov2009optimal}.  
The classical A-C framework provides a foundational decomposition of market impact into two distinct components : 
Temporary Impact: This is the immediate price concession required to execute a trade. It reflects the cost of consuming liquidity from the order book and is assumed to be ephemeral, with the price reverting once the trading pressure subsides.
Permanent Impact: This refers to the lasting shift in the asset's equilibrium price that is believed to be caused by the trade. This effect is often attributed to the information conveyed by the trading activity or a persistent alteration of the supply-demand balance \cite{cheng2017optimal}.

However, the distinction between temporary and permanent impact in the AC model, while useful for initial conceptualization, can be seen as an oversimplification. 
This empirical reality of impact  presents a direct challenge to the simplified linear impact assumption inherent in the basic Almgren-Chriss model. 
A critical theoretical constraint emerges when considering non-linear permanent market impact. Seminal work by Gueant et al.\cite{gueant2013permanent} demonstrated that within standard arbitrage-free modeling frameworks, permanent market impact must be a linear function of the instantaneous trading rate to preclude the possibility of dynamic arbitrage. Dynamic arbitrage refers to the ability to construct a sequence of trades (a round trip) that returns to the initial position while generating a positive expected profit purely by exploiting the price impact mechanism. If permanent impact were, for instance, a concave function of the trading rate alone, it could be theoretically possible to design such "money machine" strategies.
This linearity requirement for permanent impact creates a significant dilemma, as empirical observations of overall market impact often suggest non-linear, concave shapes. 
The consistent observation of concavity underscores the necessity for execution models to incorporate such non-linear effects to accurately predict and manage trading costs \cite{harford2005correlated}.
To reconcile these empirical findings with the no-arbitrage principle, several approaches have been considered. One is that the observed non-linearity resides primarily in the temporary or transient components of impact, while the true underlying permanent component remains linear \cite{brigo2019static}. 
Another avenue of research explores more sophisticated models where permanent impact might depend not only on the current trading rate but also on the cumulative volume traded so far or other path-dependent features of the execution strategy. Such modifications aim to allow for empirically consistent non-linear permanent impact characteristics while preserving the absence of arbitrage \cite{almgren2003optimal, curato2017optimal}.

The inherent dynamism of financial markets—characterized by non-linear impact, stochastic liquidity, fluctuating volatility, and the strategic behavior of participants renders static execution schedules suboptimal.
Non-linear market impact means the cost of trading varies with execution speed and size in complex ways. Stochastic liquidity implies that the capacity of the market to absorb trades is unpredictable. Trading activity introduces a reactive environment. Therefore, strategies that can adapt to these real-time changes offer the potential for significant performance improvements. 

Being aware of this, studies attempt to use machine learning and reinforecement learning in optimal execution.
RL agents learn optimal trading policies through direct interaction with a market environment (which can be a real market, a historical replay, or a sophisticated simulator) and receiving feedback in the form of rewards or penalties based on their execution quality (e.g., minimizing implementation shortfall) \cite{bergault2025hedge,hafsi2024optimal,macri2025reinforcement}.
Q-learning, Deep Q-Networks (DQN), and Double Deep Q-Networks (DDQN), are particularly well-suited for problems with complex state spaces and unknown or hard-to-model dynamics. They can implicitly learn to handle factors like time-varying liquidity and potentially non-linear impacts if the training environment accurately reflects these phenomena. Empirical studies and simulations often benchmark RL agents against traditional strategies like TWAP, VWAP, or static A-C, frequently demonstrating performance improvements. For instance, \cite{hendricks2014reinforcement} used Q-learning to dynamically adjust an A-C trajectory based on real-time spread and volume, achieving an improvement in implementation shortfall of up to 10.3\%. Similarly, DDQN has shown strong performance in environments with time-varying liquidity, outperforming analytical approximations. 
The evolution from static models to adaptive strategies reflects a fundamental trade-off between model explicitness and learning capacity.
While RL agents often demonstrate impressive performance and can master complex strategies within the specific environment they are trained on, a critical challenge arises when deploying them in live, evolving financial markets. The financial environment is often unpredictable changes, making the robustness of RL agents in new or altered market conditions typically weak, as policies optimized for previous states may not generalize effectively and can rapidly become suboptimal.

For Optimal Execution problems, IL can learn effective execution strategies by leveraging pre-existing expert demonstrations from skilled practitioners or outputs from established analytical solvers. Furthermore, IL can be particularly beneficial when explicitly defining a comprehensive reward function for RL is intricate, as the expert's actions implicitly encapsulate nuanced objectives and risk considerations.

\section{Optimal Execution Problem}

Optimal execution addresses the challenge faced by financial institutions when they need to implement a decision to buy or sell a significant quantity, $Q$, of an asset in a period. The central question is how this decision can be implemented in the market such that the desired quantity is executed at the best possible price, considering a multitude of factors including costs, risks, and market conditions. This problem is particularly acute for large orders, often termed "metaorders", which cannot be executed instantaneously without causing significant adverse price movements. The sheer size of these orders means that trading must typically occur over an extended investment horizon, $T$, which can sometimes be quite long.

The optimal execution problem seeks to liquidate $X$ shares over a finite time horizon $T$ by dividing the interval $[0, T]$ into $N$ discrete time steps. Let $t_k = k \cdot \tau$ for $k = 0, 1, \ldots, N$ where $\tau = T/N$ represents the length of each time interval. We define $n_k$ as the number of shares sold during the time interval $[t_{k-1}, t_k]$ for $k = 1, 2, \ldots, N$. The number of shares remaining at time $t_k$ is given by $x_k = X - \sum_{i=1}^{k} n_i$, with initial condition $x_0 = X$ and terminal condition $x_N = 0$. Each transaction incurs a price $p_k$ for the $n_k$ shares sold in the interval $[t_{k-1}, t_k]$. The objective is to determine the optimal trading strategy $\{n_1, n_2, \ldots, n_N\}$ that minimizes the expected execution costs while controlling for the risk associated with price uncertainty, subject to the constraint that all shares must be liquidated by time $T$.

The core challenge in optimal execution lies in navigating a fundamental set of trade-offs. These trade-offs arise because actions taken to optimize one dimension of execution quality often have repercussions for others.
There is an inherent tension between the speed of execution and the resulting market impact. Executing a large order very quickly (i.e., a high participation rate in the market) generally requires consuming liquidity aggressively. This aggressive consumption tends to push prices further, leading to higher market impact costs and a worse average execution price. Conversely, spreading the execution over a longer period by trading more slowly and passively can reduce market impact. However, this slower approach introduces other risks.

While reducing market impact by trading slowly can be beneficial from a cost perspective, it simultaneously increases the exposure to timing risk—the risk of adverse price movements occurring during the extended execution period. If the market moves significantly against the trader while the order is being patiently worked, the benefits of lower impact might be negated or even outweighed by the unfavorable price drift. Furthermore, an overly slow execution might lead to opportunity costs if the original motivation for the trade diminishes or if the execution horizon extends beyond the prediction horizon of the underlying investment strategy. Advanced optimal execution methodologies, therefore, strive to be adaptive, continuously re-evaluating and adjusting the approach to maintain an optimal balance in response to these changing factors.

This complex interplay of costs and risks gives rise to a concept analogous to the efficient frontier in portfolio theory. For a given level of execution risk (e.g., variance of execution costs), an optimal strategy would aim to minimize the expected execution cost. 

Consider a trader who needs to liquidate $X_0$ shares over a time horizon $[0,T]$. Let $X_t$ denote the remaining inventory at time $t$. The inventory dynamics are:

\begin{equation}
dX_t = -\nu_t dt, \quad X_0 = \text{initial inventory}
\end{equation}

The trader's objective is to minimize the expected execution cost plus a risk penalty:

\begin{equation}
\min_{\{\nu_t\}} \mathbb{E}\left[ \int_0^T \nu_t \tilde{S}_t dt + \lambda \text{Var}\left(\int_0^T \nu_t \tilde{S}_t dt\right) \right]
\end{equation}

subject to the constraint $X_T = 0$ (complete liquidation).

\section{Imitation Learning}
\label{sec:imitation learning}
Flow matching is typically used in imitation learning. 
However, while the flow matching policy can accurately model the desired conditional distribution $p(A_{t+1}|O_t)$, the iterative nature of ODE solving  with many steps $N$ might be too slow for the stringent low-latency requirements of optimal execution (OE). To address this, we choose Shortcut policy, which is the core for flowOE \cite{frans2024one}, denoted $s_{\phi}(a, t, \Delta t | O_t)$, which is specifically trained to generate high-quality action sequences in significantly fewer discretization steps. This policy is conditioned not only on the state $a$, time $t$, and observation $O_t$, but also on the intended integration step size $\Delta t$.

The training of this Shortcut policy $s_{\phi}$ (Algorithm~\ref{alg:Shortcut_training}, where $s_\theta$ in the algorithm represents $s_{\phi}$ with parameters $\phi$) is designed to enable it to take larger, more informed discrete steps. The procedure combines two key strategies, leveraging insights from rectified flow and consistency models \cite{song2023consistency}.

Firstly, for a portion of the training data (Algorithm~\ref{alg:Shortcut_training}, lines 4-7), the policy is guided by the direct target vector $x_1 - x_0$ (analogous to $a_E - a_0$ in our action space context). Here, $x_0 \sim \mathcal{N}(0, I)$ is a noise sample, $x_1 \sim D$ is an expert sample, and $x_t$ is an interpolation. By setting $d=0$ (line 7) for these elements, the model $s_{\phi}(x_t, t, 0 | O_t)$ is implicitly trained to align with this direct path velocity, encouraging straighter, more efficient generation paths.

Secondly, for other batch elements (Algorithm~\ref{alg:Shortcut_training}, lines 9-15), the policy is trained using a self-consistency objective inspired by methods like progressive distillation. The model $s_{\phi}(x_t, t, 2\Delta t_{\text{small}} | O_t)$ is trained to predict a velocity whose application over a larger step $2\Delta t_{\text{small}}$ matches the outcome of two consecutive smaller steps, each of size $\Delta t_{\text{small}}$. The target for this larger step is derived from stop-graded predictions of $s_{\phi}$ for these smaller constituent steps: $s_{\text{target}} \leftarrow \text{stopgrad}(s_t + s_{t+d}) / 2$, where $s_t$ and $s_{t+d}$ are velocities for the small steps and $d$ here represents $\Delta t_{\text{small}}$. This bootstrapping mechanism enables the policy to learn how to make larger, yet accurate, discrete jumps. The parameter $(d,t)$ in Algorithm~\ref{alg:Shortcut_training} (line 3) allows sampling various small step sizes $d$ and interpolation times $t$ for this consistency training.

\begin{algorithm}
\caption{ Shortcut Policy Training for OE}
\label{alg:Shortcut_training}
\begin{algorithmic}[1]
\STATE \textbf{Input:} Expert Dataset $E$, model $s_{\theta}$, batch size $B$, parameter $p$, market observations dataset $O$
\REPEAT
    \STATE Sample batch $\{x_1^i\}_{i=1}^B \sim E$, $\{x_0^i\}_{i=1}^B \sim \mathcal{N}(0, I)$, $\{(d^i, t^i)\}_{i=1}^B \sim p(d, t)$, $\{O_t^i\}_{i=1}^B \sim O$
    \FOR{$i = 1$ to $\lfloor p \cdot B \rfloor$}
        \STATE $x_t^i \leftarrow (1 - t^i) \cdot x_0^i + t^i \cdot x_1^i$
        \STATE $s_{\text{target}}^i \leftarrow (x_1^i - x_0^i)$
        \STATE $d^i \leftarrow 0$
    \ENDFOR
    \FOR{$i = \lfloor p \cdot B \rfloor + 1$ to $B$}
        \STATE $x_t^i \leftarrow (1 - t^i) \cdot x_0^i + t^i \cdot x_1^i$
        \STATE $s_t^i \leftarrow s_{\theta}(x_t^i, t^i, d^i | O_t^i)$ \COMMENT{Add market observation conditioning}
        \STATE $x_{t+d}^i \leftarrow x_t^i + d^i \cdot s_t^i$ \COMMENT{Follow ODE trajectory}
        \STATE $s_{t+d}^i \leftarrow s_{\theta}(x_{t+d}^i, t^i + d^i, d^i | O_t^i)$
        \STATE $s_{\text{target}}^i \leftarrow \text{stopgrad}((s_t^i + s_{t+d}^i) / 2)$ \COMMENT{Modified stopgrad position}
    \ENDFOR
    \STATE Update $\theta$ by minimizing $\frac{1}{B} \sum_{i=1}^B \|s_{\theta}(x_t^i, t^i, 2d | O_t^i) - s_{\text{target}}^i\|^2$ \COMMENT{Use consistent step size}
\UNTIL{convergence}
\end{algorithmic}
\end{algorithm}

Once trained, this Shortcut policy $s_{\phi}$ can generate action sequences with significantly reduced computational cost using a sampling procedure like Algorithm~\ref{alg:Shortcut_inference} (where $s_\theta$ in the algorithm is the trained $s_{\phi}$). The number of steps $M$ in Algorithm~\ref{alg:Shortcut_inference} can now be set to a small integer (e.g., 1 to 5), drastically reducing inference latency while aiming to maintain high generation quality due to the specialized training of $s_{\phi}$.

\begin{algorithm}
\caption{Shortcut Policy Inference for OE}
\label{alg:Shortcut_inference}
\begin{algorithmic}[1]
\STATE \textbf{Input:} Market observation $O_t$, trained model $s_{\theta}$, steps $M$
\STATE \textbf{Output:} Trading action sequence $a_M$
\STATE Sample $a_0 \sim \mathcal{N}(0, I)$
\STATE $d \leftarrow 1 / M$ \COMMENT{Keep step size consistent with training}
\FOR{$k = 0$ to $M-1$}
    \STATE $t_k \leftarrow k \cdot d$
    \STATE $a_{k+1} \leftarrow a_k + d \cdot s_{\theta}(a_k, t_k, d | O_t)$ \COMMENT{Use same step size d}
\ENDFOR
\RETURN $a_M$ as the generated trading action sequence
\end{algorithmic}
\end{algorithm}

\section{Data of Imitation Learning}
This section details the data generation procedure for our imitation learning framework. The process begins with the simulation of stock price movements, creating a dynamic yet controlled market environment for our experiments. Subsequently, we employ a variety of distinct expert models, each designed to execute optimal trading strategies based on these simulated price dynamics. As these expert models analyze the environment and make decisions, we systematically collect the sequences of encountered market states and the corresponding actions they take, thereby curating a comprehensive dataset of state-action pairs essential for training our imitation learning agent.

\subsection{Market Simulation}
Typically, stock price movements are described using Geometric Brownian Motion, which decomposes price evolution into two main components: a ``drift'' term to determine the underlying trend and a ``random'' term to represent unpredictable fluctuations around this trend. A critical simplifying assumption in this traditional approach is that market volatility remains constant over time. However, recognizing that real-world financial volatility is often dynamic and not fixed, our study employs the Heston model  \cite{mikhailov2004heston} to better analyze market behavior. The Heston model incorporates a stochastic process for volatility itself, allowing it to change over time and thus providing a more realistic foundation for scenarios where volatility dynamics are a key consideration.

\subsubsection{Heston Volatility}
Stochastic volatility refers to that the volatility of an asset's price is not a fixed parameter but rather a random variable that evolves over time according to its own stochastic process.
Real financial markets are characterized by volatility that is not constant but changes over time, often unpredictably. Similarly, liquidity conditions can shift. 
Additionally, empirical evidence overwhelmingly shows that volatility is not constant but exhibits clustering and mean-reversion.
As a result, market volatility is a key input to any risk-averse OE framework, as it drives the ``timing risk'' component of the execution cost \cite{chan2015optimal}.

The Heston model assumes that the instantaneous variance of the asset price follows a mean-reverting square-root process (specifically, the Cox-Ingersoll-Ross, or CIR, process). This allows volatility to fluctuate randomly over time but also to be pulled towards a long-term average level.
The Heston model is formally defined by a pair of correlated stochastic differential equations (SDEs). One SDE governs the evolution of the underlying asset price, while the other describes the dynamics of its instantaneous variance.

The SDE for the asset price, $S_t$, is given by:
\begin{equation}
dS_t = \mu S_t dt + \sqrt{\nu_t} S_t dW_t^S
\end{equation}

This equation specifies that the change in the asset price ($dS_t$) over an infinitesimal time interval $dt$ consists of two components.
The first component is a deterministic drift component, $\mu S_t dt$, where $\mu$ represents the expected instantaneous rate of return on the asset under the real-world probability measure. When pricing options, this SDE is typically considered under a risk-neutral measure, in which case $\mu$ is replaced by $r$, the risk-free interest rate (or $r - q$ if the asset pays a continuous dividend yield $q$).

The second component is a stochastic diffusion component, $\sqrt{\nu_t} S_t dW_t^S$, which captures the random fluctuations in the asset price. Crucially, the magnitude of these random movements is proportional to the current stock price $S_t$ and the square root of the instantaneous variance, $\sqrt{\nu_t}$. The term $dW_t^S$ is an increment of a standard Wiener process (also known as Brownian motion), representing the source of randomness for the asset price.

The SDE for the instantaneous variance, $\nu_t$ (which is the square of the volatility, $\sigma_t^2 = \nu_t$), is given by the Cox-Ingersoll-Ross (CIR) process, also known as a Feller square-root process:
\begin{equation}
d\nu_t = \kappa(\theta - \nu_t)dt + \xi\sqrt{\nu_t} dW_t^\nu
\end{equation}

This equation describes how the variance itself evolves stochastically.
The first is a deterministic mean-reversion component, $\kappa(\theta - \nu_t)dt$. This term pulls the current variance level $\nu_t$ back towards a long-run mean variance level $\theta$, at a speed determined by the parameter $\kappa$. If $\nu_t > \theta$, the drift is negative, pushing variance down; if $\nu_t < \theta$, the drift is positive, pushing variance up.

The second is a stochastic diffusion component, $\xi\sqrt{\nu_t} dW_t^\nu$, which introduces random fluctuations into the variance process. The magnitude of these fluctuations depends on the current level of variance ($\sqrt{\nu_t}$) and a parameter $\xi$, known as the ``volatility of volatility. The term $dW_t^\nu$ is an increment of another standard Wiener process, which is the source of randomness for the variance. The square-root term $\sqrt{\nu_t}$ is critical as it helps to ensure that the variance remains non-negative, a necessary condition for any variance model.

The two Wiener processes, $dW_t^S$ and $dW_t^\nu$, are assumed to be correlated:
\begin{equation}
\text{Corr}(dW_t^S, dW_t^\nu) = \rho \, dt
\end{equation}

The parameter $\rho$ is the instantaneous correlation coefficient between the random shocks affecting the asset price and the random shocks affecting its variance. This correlation is a key feature of the Heston model, allowing it to capture the empirically observed leverage effect, where asset prices and volatility often move in opposite directions.

\subsubsection{Non-linear Market Impact}
The assumption of linear market impact, while convenient for modeling, often falls short of capturing the true complexities observed in financial markets. 
A substantial body of empirical research compellingly suggests that instantaneous market impact is frequently a strongly concave function of trading volume. This implies that as the size of a trade increases, the additional price impact incurred for each incremental share tends to decrease. This observation stands in stark contrast to the linear assumption where each share contributes equally to the price movement.

However, a more recent and arguably more nuanced perspective introduces the concept of transient market impact \cite{gueant2013permanent,donnelly2022optimal,curato2017optimal}. Instead, the impact is seen to gradually decay over time as the market absorbs the trade and liquidity is replenished in the order book. This framework inherently models the resilience of the order book—its ability to recover from liquidity shocks.

When executing a trade at rate $\nu_t$, the trader faces both temporary and permanent market impacts. The execution price $\tilde{S}_t$ differs from the mid-market price due to temporary impact:

\begin{equation}
\tilde{S}_t = S_t - f(\nu_t)
\end{equation}

where $f(\cdot)$ is the temporary impact function. We adopt a non-linear power-law specification:

\begin{equation}
f(\nu) = \epsilon \cdot |\nu|^{\beta}
\end{equation}

where:
$\epsilon > 0$ is the temporary impact coefficient
$\beta \in (0, 1]$ is the non-linearity parameter

The parameter $\beta$ captures different market microstructure effects:
$\beta = 1$: Linear impact (traditional Almgren-Chriss model)
$\beta = 0.5$: Square-root impact (empirically observed in many markets)
$\beta < 1$: Concave impact function (economies of scale in execution)

\subsection{Data Collection}
To establish a set of expert benchmarks for our study, we prepared and utilized several distinct approaches. These include a range of traditional optimal execution models, known for their analytical tractability, as well as more reinforcement learning methods designed to adapt to market dynamics. These models collectively serve as the ``experts" in our framework, and a detailed description of each, including their underlying principles and specific configurations, is provided in  \ref{appendix:expertmodel}.

We collect comprehensive trajectory data including:
State vectors: $(t, X_t, S_t, V_t)$
Actions: $\nu_t$
Market parameters: $(\mu, \kappa, \theta, \xi, \rho, \eta, \epsilon, \beta)$

\paragraph{Simulation Environment}
The environment simulates the liquidation of $X_0 = 10,000$ shares over a trading horizon of $T = 1.0$ (normalized to one trading day), discretized into $N = 100$ intervals, corresponding to approximately 4-minute steps in a 6.5-hour trading day. The initial stock price is set to $S_0 = 100$. Price and variance dynamics follow the discretized Heston model, with execution prices computed as:
\begin{equation}
\tilde{S}_k = S(t_k) - \epsilon \nu_k^\beta, \label{eq:execution_price_discrete}
\end{equation}
where $\nu_k = x_k / \Delta t$ is the trading rate, $x_k$ is the shares traded at step $k$, $\epsilon$ is the temporary impact coefficient, and $\beta$ is the non-linearity parameter.

\paragraph{Parameter Space}
We explore a comprehensive parameter space to assess strategy robustness across diverse market conditions. Table~\ref{tab:parameters} summarizes the parameters, selected based on empirical evidence from equity markets.

\begin{table}[ht]
\centering
\caption{Simulation Parameter Values}
\begin{tabular}{llc}
\toprule
\textbf{Parameter} & \textbf{Symbol} & \textbf{Values} \\
\midrule
\multicolumn{3}{l}{\textit{Market Dynamics}} \\
Drift rate & $\mu$ & $\{0.0, 0.02\}$ \\
Initial volatility & $\sqrt{V_0}$ & $\{0.2, 0.3, 0.4\}$ \\
Mean reversion speed & $\kappa$ & $2.0$ \\
Long-term volatility & $\sqrt{\theta}$ & $0.3$ \\
Volatility of variance & $\xi$ & $\{0.2, 0.3\}$ \\
Correlation & $\rho$ & $-0.7$ \\
\midrule
\multicolumn{3}{l}{\textit{Market Impact}} \\
Permanent impact & $\eta$ & $2.5 \times 10^{-5}$ \\
Temporary impact coefficient & $\epsilon$ & $5.0 \times 10^{-5}$ \\
Non-linearity parameter & $\beta$ & $\{0.2, 0.6, 0.9\}$ \\
\midrule
\multicolumn{3}{l}{\textit{Risk Preferences}} \\
Risk aversion & $\lambda$ & $10^{-5}$ \\
\bottomrule
\end{tabular}
\label{tab:parameters}
\end{table}

The parameter choices are justified as follows:
For \textit{Initial volatility} ($\sqrt{V_0} \in \{0.2, 0.3, 0.4\}$) spans normal to stressed market conditions, reflecting annual volatilities of 20\% to 40\%.
For \textit{Mean reversion speed} ($\kappa = 2.0$) implies a volatility shock half-life of approximately $\ln(2)/\kappa \approx 0.347$ years (4.16 months), consistent with equity market dynamics .
For \textit{Correlation} ($\rho = -0.7$) captures the leverage effect, where negative price shocks correlate with increased volatility.
For \textit{Non-linearity parameter} ($\beta \in \{0.5, 1.0, 1.5\}$) covers square-root impact ($\beta = 0.5$, linear impact ($\beta = 1.0$), and super-linear impact ($\beta = 1.5$) to test strategy sensitivity.
For \textit{Market impact coefficients} ($\eta, \epsilon$) are calibrated to typical equity market values, ensuring realistic price impacts for large orders.

For each combination of parameters ($\mu$, $\sqrt{V_0}$, $\xi$, $\beta$), we simulate $n = 100$ independent episodes to ensure statistical reliability, using a base seed of 42 for reproducibility. The total number of episodes is:
\begin{equation}
|\mu| \times |\sqrt{V_0}| \times |\xi| \times |\beta| \times |\text{Strategies}| \times n = 2 \times 3 \times 2 \times 3 \times 4 \times 100 = 14,400.
\end{equation}
Each episode generates a trajectory of 100 time steps, yielding approximately 1.44 million state-action data points. For each time step $k$, we collect:
\begin{itemize}
    \item \textit{State variables}: Time remaining $\tau_k = T - t_k$, inventory $q_k$, mid-price $S_k$, variance $V_k$.
    \item \textit{Action}: Shares executed $x_k$.
    \item \textit{Outcomes}: Execution price $\tilde{S}_k$, cash $C_k$, and implementation shortfall $IS$  at the episode’s end.
\end{itemize}

\section{Experiments and Results Analysis}
This section outlines our experimental design. Furthermore, we define the specific evaluation metrics that will be used to quantify and compare the performance of different approaches.

\subsection{Benchmark}
\textbf{TWAP}
This strategy divides the total order size by the number of time steps in the execution horizon and executes an equal number of shares at each time step.

The TWAP strategy liquidates the inventory uniformly over time:
\begin{equation}
    \nu_k = \frac{X_0}{T}, \quad x_k = \frac{X_0}{N}, \quad k = 0, 1, \dots, N-1.
\end{equation}
This strategy is simple and ignores market conditions, serving as a baseline.

\textbf{VWAP}
This strategy aims to match the volume distribution of the market. We estimate the expected volume at each time step based on historical data and execute a proportional number of shares.

The VWAP strategy aligns trading with an assumed volume curve, modeled as a quadratic function:
\begin{equation}
    \omega_k = \frac{2.5 \left( \frac{k}{N} - 0.5 \right)^2 + 0.5}{\sum_{j=0}^{N-1} \left[ 2.5 \left( \frac{j}{N} - 0.5 \right)^2 + 0.5 \right]},
\end{equation}
where $\omega_k$ is the normalized volume fraction at step $k$. The shares traded are:
\begin{equation}
    x_k = X_0 \omega_k, \quad \nu_k = \frac{x_k}{\Delta t}.
\end{equation}

\textbf{Almgren-Chriss Approximation (AC-Approx)}
The AC-Approx strategy adapts the Almgren-Chriss model to the non-linear impact setting by approximating the temporary impact as an effective linear coefficient:
\begin{equation}
    \epsilon_{\text{eff}} = \epsilon \left( \frac{X_0}{T} \right)^{\beta - 1}.
\end{equation}
The trading rate is derived from the optimal inventory trajectory:
\begin{equation}
    q(t) = X_0 \frac{\sinh \left( \kappa (T - t) \right)}{\sinh \left( \kappa T \right)}, \quad \kappa = \sqrt{\frac{\lambda \theta}{\epsilon_{\text{eff}}}},
\end{equation}
where $\theta$ is the long-term variance, and $\lambda$ is the risk aversion parameter. The shares traded at step $k$ are:
\begin{equation}
    x_k = q(t_k) - q(t_{k+1}), \quad \nu_k = \frac{x_k}{\Delta t}.
\end{equation}
If $\kappa T$ is small, the strategy reverts to TWAP-like behavior:
\begin{equation}
    x_k \approx \frac{X_0}{N}.
\end{equation}

\textbf{Heston-Optimal Strategy}
The Heston-Optimal strategy accounts for stochastic volatility and non-linear impact, using an analytical trading rate derived from the optimal control problem:
\begin{equation}
    \nu^*(t) = (1 + \beta) \frac{q(t)}{T - t} \left( \frac{\sqrt{\theta + (V(t) - \theta) e^{-\kappa (T - t)}}}{\sqrt{V(t)}} \right)^{1/2},
\end{equation}
where $q(t)$ is the current inventory, $V(t)$ is the current variance, and $T - t$ is the time remaining. For numerical stability near $t \approx T$, a smoothed transition to uniform liquidation is applied:
\begin{equation}
    \nu^*(t) = \begin{cases} 
        \alpha \nu^*(t) + (1 - \alpha) \frac{q(t)}{T - t}, & \text{if } T - t < 0.05 T, \\
        \nu^*(t), & \text{otherwise},
    \end{cases}
\end{equation}
where $\alpha = (T - t)/(0.05 T)$. The shares traded are:
\begin{equation}
    x_k = \nu_k^* \Delta t, \quad \nu_k^* = \nu^*(t_k).
\end{equation}
In the final step ($k = N-1$), all remaining inventory is liquidated: $x_{N-1} = q(t_{N-1})$.

\subsection{Evaluation Metrics}
We evaluate the performance of the different execution strategies using the following metrics:

\textbf{AC (Almgren-Chriss Objective):} Expected implementation shortfall plus a risk penalty proportional to its variance, $\mathbb{E}[IS] + \lambda \text{Var}[IS]$, where $\lambda$ is the risk aversion parameter (\textit{lower is better})

\textbf{Standard Deviation:} Volatility of implementation shortfall

\textbf{Volatility of Execution Cost:} The standard deviation of the execution cost across multiple simulations. This measures the risk associated with the execution strategy.

\textbf{Implied Shortfall:} It is a pre-trade estimate of the total expected execution cost for a large order, predicting the difference between a benchmark and the anticipated average execution price using models for costs such as market impact and spreads.

The implementation shortfall (IS) measures the difference between the ideal execution value and the actual execution proceeds:

\begin{equation}
\text{IS} = X_0 S_0 - \int_0^T \nu_t \tilde{S}_t dt
\end{equation}

To evaluate the performance of different strategies, we employ a Monte Carlo simulation approach:

\begin{enumerate}
    \item Generate $N$ sample paths of $(S_t, V_t)$ using the Milstein scheme for improved accuracy
    \item For each path and strategy, compute the trading trajectory $\{\nu_t\}$
    \item Calculate the implementation shortfall for each simulation
    \item Compute performance metrics: mean IS, standard deviation, and the Almgren-Chriss objective $\mathbb{E}[\text{IS}] + \lambda\text{Var}[\text{IS}]$
\end{enumerate}

The discretized dynamics for numerical implementation are:

\begin{align}
\log S_{t+\Delta t} &= \log S_t + \left(\mu - \frac{V_t}{2} - \eta\nu_t\right)\Delta t + \sqrt{V_t\Delta t} Z_1 \\
V_{t+\Delta t} &= V_t + \kappa(\theta - V_t)\Delta t + \xi\sqrt{V_t\Delta t}(\rho Z_1 + \sqrt{1-\rho^2}Z_2) \\
&\quad + \frac{\xi^2\Delta t}{4}((\rho Z_1 + \sqrt{1-\rho^2}Z_2)^2 - 1) \nonumber
\end{align}

where $Z_1, Z_2 \sim \mathcal{N}(0,1)$ are independent standard normal random variables.

\subsection{Results Analysis}
To empirically validate the efficacy of our proposed FlowOE framework, we conducted a series of comprehensive experiments across several  market conditions, each designed to reflect distinct real-world scenarios characterized by varying levels of volatility and market impact.
Traditional studies often cite market impact parameters in the range of 0.4 to 0.7 to characterize this concavity, our study comprehensively covers these cases. 

The empirical results clearly demonstrate the superior performance of the Shortcut Model across all tested market conditions and evaluation metrics. 
`Shortcut Model HO' = Shortcut policy trained on Heston-Optimal expert demonstrations.
`Shortcut Model PPO' = Shortcut policy trained on PPO expert demonstrations.
Notably, Shortcut consistently achieved significantly lower Implied Shortfall (IS), reduced risk (STD), and a substantially improved overall cost function (AC) when compared to the AC-Approx expert model, indicating that it not only successfully learned from the expert but also effectively refined those strategies for enhanced outcomes. Furthermore, Shortcut substantially outperformed standard industry benchmarks TWAP and VWAP, as well as the Heston-Optimal and its predecessor, the Shortcut Model, highlighting its practical utility and robustness in diverse market scenarios involving varying levels of volatility and market impact. This consistent best-in-class performance underscores the efficacy of the Shortcut model's design for advanced optimal execution.

Scenario Definitions are the following:

\textbf{HH (High Volatility \& High Impact):} High market volatility (40\%) with significant market impact costs

\textbf{HL (High Volatility \& Low Impact):} High market volatility (40\%) with reduced market impact costs  

\textbf{LH (Low Volatility \& High Impact):} Low market volatility (20\%) with significant market impact costs

\textbf{LL (Low Volatility \& Low Impact):} Low market volatility (20\%) with reduced market impact costs

\begin{table}[!htbp]
  \centering
  \caption{Performance Metrics for Different Strategies Across Market Conditions, Results are for $\beta=0.5$ (square-root impact model, concave market). Metrics are evaluated out-of-sample over 10000 trials for significance and convergence.}
  \renewcommand{\arraystretch}{1.2}
  \resizebox{\textwidth}{!}{%
  \begin{tabular}{lrrr|rrr|rrr|rrr}
  \toprule
  & \multicolumn{3}{c|}{High Vol. \& High Impact (HH)} & \multicolumn{3}{c|}{High Vol. \& Low Impact (HL)} & \multicolumn{3}{c|}{Low Vol. \& High Impact (LH)} & \multicolumn{3}{c}{Low Vol. \& Low Impact (LL)} \\
  & IS $\downarrow$ & STD $\downarrow$ & AC $\downarrow$ & IS $\downarrow$ & STD $\downarrow$ & AC $\downarrow$ & IS $\downarrow$ & STD $\downarrow$ & AC $\downarrow$ & IS $\downarrow$ & STD $\downarrow$ & AC $\downarrow$ \\
  \midrule
  TWAP & 211138.91 & 162927.23 & 476591.73 & 47917.66 & 206859.54 & 475826.36 & 211142.58 & 82956.37 & 279960.17 & 47886.68 & 105391.33 & 158960.00 \\
  VWAP & 210907.97 & 156429.74 & 455610.61 & 47776.54 & 200470.78 & 449661.89 & 210953.62 & 79659.98 & 274410.75 & 47796.42 & 102156.86 & 152156.65 \\
  AC-Approx & 210879.60 & 147140.73 & 427383.55 & 47251.45 & 148179.80 & 266823.99 & 211104.94 & 80642.05 & 276136.34 & 47781.70 & 93711.19 & 135599.58 \\
  Heston-Optimal & 210591.53 & 140005.39 & 406606.62 & 47672.24 & 178424.05 & 366023.67 & 210712.19 & 71358.61 & 261632.69 & 47742.15 & 91010.36 & 130571.01 \\
  Shortcut Model HO & 210994.31 & 142105.83 & 412934.98 & 49260.87 & 179651.98 & 372009.22 & 209749.75 & 65922.60 & 253207.64 & 48184.77 & 83807.90 & 118422.41 \\
  Shortcut Model PPO &  189332.04 &  89339.09&  269146.78&  43336.09&  105978.40&  155650.29&  185448.12&  82749.77&  253923.36&  41623.71&  94217.13&  130392.40\\

  \bottomrule
  \end{tabular}%
  }
  \label{tab:performance_metrics_beta05_corrected}
\end{table}

\begin{table}[!htbp]
  \centering
  \caption{Performance Metrics for Different Strategies Across Market Conditions, Results are for $\beta=0.8$ (square-root impact model, less concave market). Metrics are evaluated out-of-sample over 10000 trials for significance and convergence.}
  \renewcommand{\arraystretch}{1.2}
  \resizebox{\textwidth}{!}{%
  \begin{tabular}{lrrr|rrr|rrr|rrr}
  \toprule
  & \multicolumn{3}{c|}{High Vol. \& High Impact (HH)} & \multicolumn{3}{c|}{High Vol. \& Low Impact (HL)} & \multicolumn{3}{c|}{Low Vol. \& High Impact (LH)} & \multicolumn{3}{c}{Low Vol. \& Low Impact (LL)} \\
  & IS $\downarrow$ & STD $\downarrow$ & AC $\downarrow$ & IS $\downarrow$ & STD $\downarrow$ & AC $\downarrow$ & IS $\downarrow$ & STD $\downarrow$ & AC $\downarrow$ & IS $\downarrow$ & STD $\downarrow$ & AC $\downarrow$ \\
  \midrule
  TWAP & 212623.80 & 162927.23 & 478076.63 & 48214.64 & 206859.54 & 476123.34 & 212627.47 & 82956.37 & 281445.06 & 48183.66 & 105391.33 & 159256.98 \\
  VWAP & 212470.74 & 156429.74 & 457173.39 & 48089.10 & 200470.78 & 449974.45 & 212516.40 & 79659.98 & 275973.52 & 48108.97 & 102156.86 & 152469.21 \\
  AC-Approx & 212611.05 & 161751.97 & 474248.04 & 48167.50 & 200230.36 & 449089.46 & 212625.43 & 82803.72 & 281190.00 & 48178.01 & 104502.13 & 157384.96 \\
  Heston-Optimal & 212029.51 & 130390.69 & 382046.83 & 47872.33 & 166482.93 & 325037.99 & 212185.06 & 66364.04 & 256226.92 & 47994.23 & 84807.63 & 119917.57 \\
  Shortcut Model HO& 213409.04 & 142074.08 & 415259.48 & 49744.36 & 179647.47 & 372476.51 & 212304.58 & 65912.20 & 255748.76 & 48699.40 & 83807.73 & 118936.75 \\
  Shortcut Model PPO &  161140.96 &  40110.31&  177229.32&  33168.93&  28701.35&  41406.60&  217900.34&  42911.147&  236313.99&  45353.33&  38667.25&  60304.89\\
  
  \bottomrule
  \end{tabular}%
  }
  \label{tab:performance_metrics_beta08_corrected}
\end{table}

\begin{table}[!htbp]
  \centering
  \caption{Performance Metrics for Different Strategies Across Market Conditions, Results are for $\beta=0.3$ (square-root impact model, highly concave market). Metrics are evaluated out-of-sample over 10000 trials for significance and convergence.}
  \renewcommand{\arraystretch}{1.2}
  \resizebox{\textwidth}{!}{%
  \begin{tabular}{lrrr|rrr|rrr|rrr}
  \toprule
  & \multicolumn{3}{c|}{High Vol. \& High Impact (HH)} & \multicolumn{3}{c|}{High Vol. \& Low Impact (HL)} & \multicolumn{3}{c|}{Low Vol. \& High Impact (LH)} & \multicolumn{3}{c}{Low Vol. \& Low Impact (LL)} \\
  & IS $\downarrow$ & STD $\downarrow$ & AC $\downarrow$ & IS $\downarrow$ & STD $\downarrow$ & AC $\downarrow$ & IS $\downarrow$ & STD $\downarrow$ & AC $\downarrow$ & IS $\downarrow$ & STD $\downarrow$ & AC $\downarrow$ \\
  \midrule
  TWAP & 211054.76 & 162927.23 & 476507.58 & 47900.83 & 206859.54 & 475809.53 & 211058.42 & 82956.37 & 279876.01 & 47869.85 & 105391.33 & 158943.17 \\
  VWAP & 210821.42 & 156429.74 & 455524.06 & 47759.23 & 200470.78 & 449644.58 & 210867.07 & 79659.98 & 274324.20 & 47779.10 & 102156.86 & 152139.34 \\
  AC-Approx & 209153.11 & 109162.99 & 328318.70 & 45793.91 & 94434.47 & 134972.61 & 210706.75 & 71236.80 & 261453.56 & 47221.72 & 67760.45 & 93136.51 \\
  Heston-Optimal & 210734.65 & 147694.01 & 428869.86 & 47759.28 & 187963.32 & 401061.38 & 210818.50 & 75347.65 & 267591.18 & 47790.11 & 95965.76 & 139884.38 \\
  Shortcut Model HO& 210882.39 & 142106.79 & 412825.78 & 49238.46 & 179652.13 & 371987.34 & 209633.80 & 65922.95 & 253092.15 & 48161.46 & 83807.92 & 118399.13 \\
  Shortcut Model PPO &  188484.20 &  34989.05&  200726.53&  38631.46&  28432.55&  46715.56&  230022.31&  38655.45&  244964.74&  47767.46&  38475.53&  62571.12\\
 
  \bottomrule
  \end{tabular}%
  }
  \label{tab:performance_metrics_beta03_corrected}
\end{table}

\begin{table}[!htbp]
  \centering
  \caption{Parameter Specifications for Market Scenarios ($\beta = 0.5$, Concave Market Impact)}
  \begin{adjustbox}{width=1.1\columnwidth,center}
  \renewcommand{\arraystretch}{1.3}
  \begin{tabular}{lcccc}
  \toprule
  \textbf{Parameter} & \textbf{HH Scenario} & \textbf{HL Scenario} & \textbf{LH Scenario} & \textbf{LL Scenario} \\
  \midrule
  \multicolumn{5}{l}{\textit{Common Parameters}} \\
  Time horizon ($T$) & 1.0 & 1.0 & 1.0 & 1.0 \\
  Initial shares ($X_0$) & 10,000 & 10,000 & 10,000 & 10,000 \\
  Initial price ($S_0$) & \$100 & \$100 & \$100 & \$100 \\
  Drift rate ($\mu$) & 0.0 & 0.0 & 0.0 & 0.0 \\
  Mean reversion speed ($\kappa$) & 2.0 & 2.0 & 2.0 & 2.0 \\
  Price-variance correlation ($\rho$) & -0.7 & -0.7 & -0.7 & -0.7 \\
  Impact non-linearity ($\beta$) & 0.5 & 0.5 & 0.5 & 0.5 \\
  Risk aversion ($\gamma$) & $1 \times 10^{-5}$ & $1 \times 10^{-5}$ & $1 \times 10^{-5}$ & $1 \times 10^{-5}$ \\
  \midrule
  \multicolumn{5}{l}{\textit{Volatility Parameters}} \\
  Initial variance ($V_0$) & 0.16 (40\% vol) & 0.16 (40\% vol) & 0.04 (20\% vol) & 0.04 (20\% vol) \\
  Long-term variance ($\theta$) & 0.16 (40\% vol) & 0.16 (40\% vol) & 0.04 (20\% vol) & 0.04 (20\% vol) \\
  Volatility of variance ($\xi$) & 0.5 & 0.5 & 0.2 & 0.2 \\
  \midrule
  \multicolumn{5}{l}{\textit{Market Impact Parameters}} \\
  Permanent impact ($\eta$) & $5 \times 10^{-5}$ & $1 \times 10^{-5}$ & $5 \times 10^{-5}$ & $1 \times 10^{-5}$ \\
  Temporary impact ($\epsilon$) & $1 \times 10^{-4}$ & $2 \times 10^{-5}$ & $1 \times 10^{-4}$ & $2 \times 10^{-5}$ \\
  \bottomrule
  \end{tabular}
  \label{tab:scenario_parameters}
  \end{adjustbox}
\end{table}

\newpage 

\begin{figure}[htbp]
    \centering
    \includegraphics[width=0.8\textwidth]{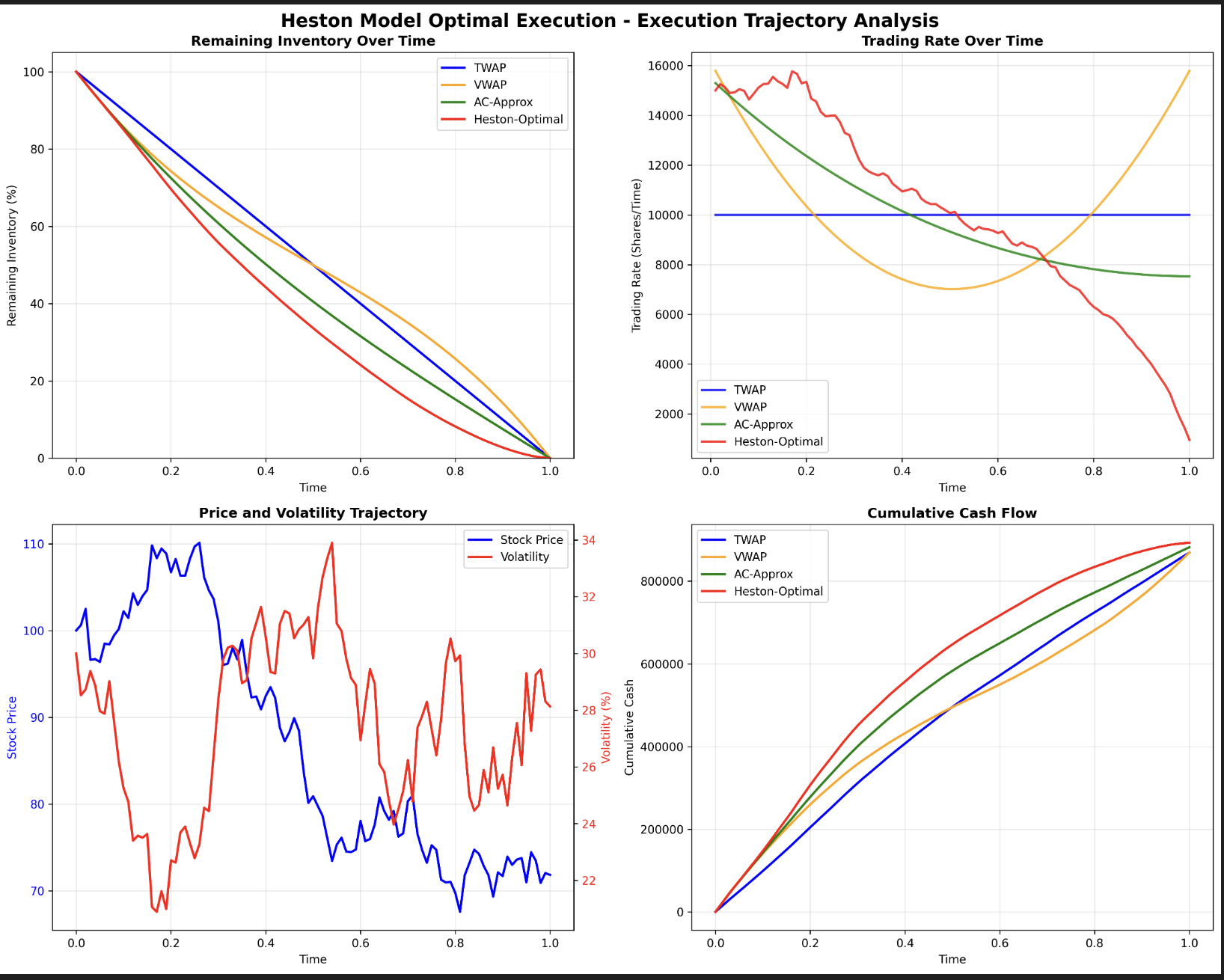}
    \caption{Single Trajectory}
    \label{fig:trajectory}
\end{figure}

\begin{figure}[htbp]
    \centering
    \includegraphics[width=0.8\textwidth]{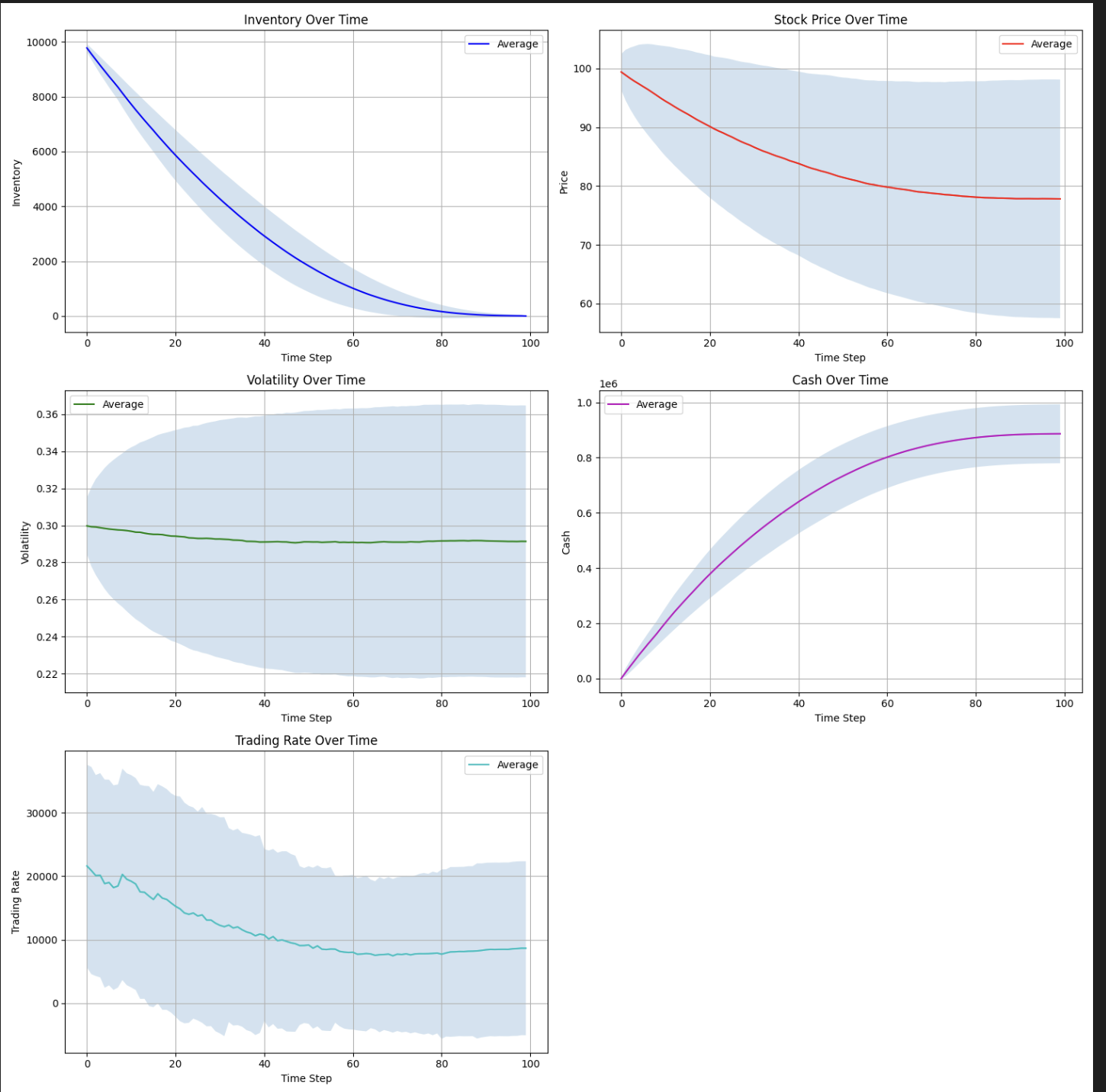}
    \caption{Shortcut Policy Learning the HO Experts}
    \label{fig:Shortcut}
\end{figure}

\begin{figure}[htbp]
    \centering
    \includegraphics[width=0.8\textwidth]{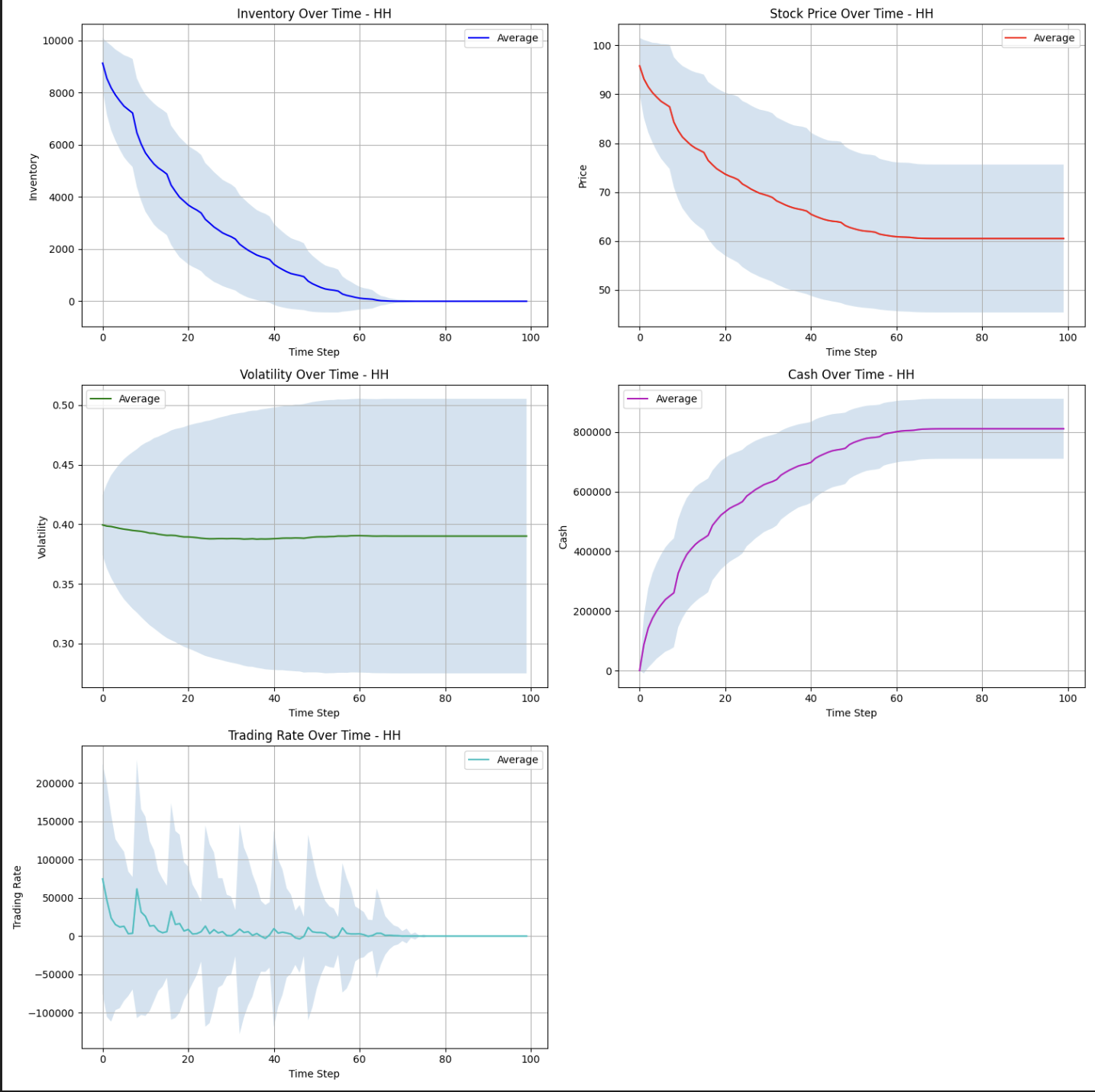}
    \caption{Shortcut Policy Learning the PPO Experts}
    \label{fig:sc_ppo}
\end{figure}

Looking at Figures \ref{fig:trajectory}, \ref{fig:Shortcut}, \ref{fig:sc_ppo}, we illustrate the systematic liquidation of a significant asset inventory over 100 time steps, leading to a consistent, smooth depletion down to near zero. 
The middle curve is the mean while the blue shade area indicates standard error.
This selling activity predictably exerted downward pressure on the stock price, demonstrating market impact, with the price declining from around 102 to below 80. Concurrently, cash reserves steadily accumulated, though the rate of accumulation slowed as the asset price fell and inventory diminished. Average market volatility remained relatively stable around 0.30. Notably, while the inventory reduction was highly consistent, the stock price path and final cash values showed increasing uncertainty (standard error) over time. The mean trading rate itself displayed a dynamic pattern—starting high, decreasing through the middle stages, and then partially recovering towards the end—and exhibited considerable variability across scenarios, suggesting either an adaptive strategy or an aggregation of diverse execution approaches.

\newpage
\section{Conclusion}
This study introduced FlowOE, a flow matching-based imitation learning framework for optimal execution in stochastic financial markets. By employing the Shortcut policy approach, we addressed computational latency constraints while maintaining powerful distributional modeling capabilities.
Our empirical evaluation across diverse market conditions demonstrated that FlowOE consistently outperformed traditional benchmarks. In highly concave markets ($\beta$ = 0.3), the framework achieved implementation shortfall reductions of up to 10.9\% compared to Almgren-Chriss approximations while reducing risk by 68\%. The Shortcut policy's ability to reduce inference from hundreds of ODE steps to just 1-5 steps makes real-time deployment feasible.
The framework's success stems from three key contributions: (1) effective learning from diverse expert strategies with adaptive behavior selection, (2) significant latency reduction through specialized training procedures, and (3) the ability to discover refinements beyond expert performance through fine-tuning.
These results have immediate practical implications for institutional trading, offering a path toward more adaptive and cost-effective execution strategies. Future research directions include extending to multi-asset problems, incorporating additional market microstructure features, and exploring integration with reinforcement learning approaches.
To the best of our knowledge, this work represents the first successful application of flow matching models to stochastic optimal execution, demonstrating both theoretical advances and empirically validated performance improvements that can enhance trading outcomes and market efficiency.

\newpage
\begin{appendix}
\section{Market Dynamics}
The mid-price $S(t)$ and variance $V(t)$ follow the Heston model with permanent market impact:
\begin{align}
dS(t) &= \left( \mu - \eta \nu(t) \right) S(t) dt + \sqrt{V(t)} S(t) dW_S(t), \label{eq:price_heston} \\
dV(t) &= \kappa \left( \theta - V(t) \right) dt + \xi \sqrt{V(t)} dW_v(t), \label{eq:variance_heston}
\end{align}
where parameters are defined in Table~\ref{tab:parameters}. The execution price includes non-linear temporary impact:
\begin{equation}
\tilde{S}(t) = S(t) - \epsilon \nu(t)^\beta, \label{eq:execution_price}
\end{equation}
with $\epsilon > 0$ and $\beta \in (0,1]$ (e.g., $\beta = 0.5$ for square-root impact).

\begin{table}[ht]
\centering
\caption{Model Parameters}
\begin{tabular}{cl}
\hline
Parameter & Description \\
\hline
$\mu$ & Drift rate of the asset price \\
$\eta$ & Permanent impact coefficient \\
$\nu(t)$ & Trading rate (shares per unit time) \\
$\kappa$ & Mean reversion speed of variance \\
$\theta$ & Long-term variance level \\
$\xi$ & Volatility of variance \\
$\rho$ & Correlation between $W_S(t)$ and $W_v(t)$ \\
$\epsilon$ & Temporary impact coefficient \\
$\beta$ & Non-linearity parameter for temporary impact \\
$\lambda$ & Risk aversion parameter \\
\hline
\end{tabular}
\label{tab:parameters}
\end{table}

\paragraph{Inventory and Cash Dynamics}
Inventory and cash evolve as:
\begin{align}
dq(t) &= -\nu(t) dt, \quad q(0) = X_0, \quad q(T) \geq 0, \label{eq:inventory} \\
dC(t) &= \nu(t) \left( S(t) - \epsilon \nu(t)^\beta \right) dt. \label{eq:cash}
\end{align}

\section{Expert Models for Optimal Execution}
\label{appendix:expertmodel}

\subsection{Almgren-Chriss model}
Consider a financial asset whose mid-price $S_t$ evolves as
\begin{equation}
\mathrm{d}S_t = \sigma\,\mathrm{d}W_t - \gamma\,v_t\,\mathrm{d}t,
\end{equation}
where $\sigma>0$ is the volatility, $\gamma>0$ is the permanent impact coefficient, and $v_t\ge0$ is the trading rate (sell rate). The inventory dynamics satisfy
\begin{equation}
\mathrm{d}q_t = -v_t\,\mathrm{d}t,\quad q_0=Q_0.
\end{equation}
Trades incur a temporary impact cost $\eta v_t^2$ per unit time and a permanent impact cost $\gamma q_t v_t$. Remaining inventory $q_T$ is settled at price $S_T$. We minimize the expected total cost
\begin{equation}
J(v)=\mathbb{E}\Bigl[\int_0^T(\eta v_t^2+\gamma q_t v_t)\,\mathrm{d}t - q_T S_T\Bigr].
\end{equation}

\paragraph{HJB Formulation}

Define the value function
\begin{equation}
V(t,S,q)=\inf_{v_{[t,T]}}\mathbb{E}\Bigl[\int_t^T(\eta v_s^2+\gamma q_s v_s)\mathrm{d}s - q_T S_T \;\Big|\;S_t=S,q_t=q\Bigr],
\end{equation}
with terminal condition $V(T,S,q)=-qS$. The HJB equation reads
\begin{equation}
0 = V_t + \frac{1}{2}\sigma^2 V_{SS} + \inf_{v\ge0}\{ -vV_q - \gamma v V_S + \eta v^2 + \gamma q v\}.
\end{equation}
The first-order condition gives
\begin{equation}
v^*=\frac{V_q+\gamma V_S-\gamma q}{2\eta}.
\end{equation}
Substitution yields the PDE
\begin{equation}
0=V_t + \frac{1}{2}\sigma^2 V_{SS} - \frac{(V_q+\gamma V_S-\gamma q)^2}{4\eta}.
\end{equation}

\paragraph{Quadratic Ansatz and ODE System} 

We postulate a quadratic form
\begin{equation}
V(t,S,q)=A(t)q^2+B(t)S q + C(t)S^2 + D(t).
\end{equation}
Matching coefficients in the HJB PDE leads to the following Riccati ODEs:
\begin{align}
A'(t)&=\frac{[2A(t)+\gamma B(t)-\gamma]^2}{4\eta},\quad A(T)=0,\\
B'(t)&=\frac{[2A(t)+\gamma B(t)-\gamma][B(t)+2\gamma C(t)]}{2\eta},\quad B(T)=-1,\\
C'(t)+\sigma^2C(t)&=\frac{[B(t)+2\gamma C(t)]^2}{4\eta},\quad C(T)=0,\\
D'(t)&+\sigma^2 C(t)=0,\quad D(T)=0.
\end{align}

\paragraph{Closed-Form Solution}
Setting $\Delta=\gamma^2+4\eta$, the solutions are
\begin{align}
A(t)&=\frac{\sqrt{\Delta}-\gamma}{4\eta}\coth\bigl(\tfrac{\sqrt{\Delta}}{2\eta}(T-t)\bigr)-\frac{\gamma}{2},\\
B(t)&=-\frac{\sqrt{\Delta}}{\sqrt{\Delta}\cosh(\sqrt{\Delta}(T-t))-\gamma\sinh(\sqrt{\Delta}(T-t))},\\
C(t)&=\frac{\eta\sinh(\sqrt{\Delta}(T-t))}{4(\sqrt{\Delta}\cosh(\sqrt{\Delta}(T-t))-\gamma\sinh(\sqrt{\Delta}(T-t))) }.
\end{align}
Thus the state-feedback control is
\begin{equation}
v^*(t,S_t,q_t)=\alpha_1(t)q_t+\alpha_2(t)S_t,
\end{equation}
with
\begin{equation}
\alpha_1(t)=\frac{2A(t)+\gamma B(t)-\gamma}{2\eta},\quad
\alpha_2(t)=\frac{B(t)+2\gamma C(t)}{2\eta}.
\end{equation}

\subsection{Expert Trajectory Generation via Proximal Policy Optimization}
\label{sec:ppo_expert_data_collection}

To generate diverse and robust expert demonstrations for training the FlowOE imitation learning framework, we employ Proximal Policy Optimization (PPO) agents \citep{schulman2017proximal}. This approach proves particularly valuable in scenarios where analytical solutions are unavailable or when greater adaptability is required. The PPO agents serve as sophisticated experts that are specifically trained for optimal execution in a simulated Heston market environment.

We train individual PPO agents that are tailored to distinct market scenarios. These scenarios are categorized by volatility levels (High/Low) and market impact characteristics (High/Low), resulting in four distinct configurations: HH, HL, LH, and LL . Each PPO agent is trained to minimize an execution cost function, which is equivalently expressed as maximizing cumulative rewards that penalize execution costs, price variance risks, and residual inventory at the terminal time $T$.

The PPO agents operate within a custom Gym-compatible environment \texttt{(`OptimalExecutionHestonEnv')} \citep{brockman2016openaigym} that simulates Heston volatility dynamics, nonlinear temporary impact, and linear permanent impact. 

At discrete time points $t_k$, the agents observe the following state variables:
\begin{itemize}
\item Remaining time until termination: $T - t_k$
\item Current inventory position: $q(t_k)$
\item Mid-price of the asset: $S(t_k)$
\item Current volatility level: $V(t_k)$
\end{itemize}

These observations are normalized prior to input into the PPO policy network, which is typically implemented as a multilayer perceptron. The network outputs the parameters (mean and standard deviation) of Gaussian action distributions. The actions represent fractions of inventory to be liquidated at each time step $\Delta t$, which are subsequently converted into discrete share quantities. A critical constraint is enforced to ensure complete liquidation of all remaining inventory in the final step, thereby guaranteeing zero terminal inventory.

Training continues until performance convergence is achieved for each scenario, at which point the optimal models are saved for subsequent use in trajectory generation.

\subsection{Systematic Generation of Expert Trajectories}

Following the training phase, the PPO expert models generate extensive state-action trajectories across systematically varied market parameters. This process enhances the comprehensiveness of the dataset and ensures robust coverage of the parameter space. The data collection methodology comprises four key stages:

\textbf{Parameter Grid Construction.} We define a comprehensive parameter grid that systematically varies the initial volatility ($\sqrt{V_0}$), volatility of variance ($\xi$), market impact nonlinearity ($\beta$), and other relevant Heston model parameters including the drift ($\mu$), mean reversion rate ($\kappa$), long-term variance ($\theta$), correlation coefficient ($\rho$), and impact parameters ($\eta$, $\epsilon$).

\textbf{Scenario-Specific Model Selection.} Each parameter combination is mapped to one of the four scenarios (HH, HL, LH, LL) based on its volatility and market impact characteristics. The appropriate pre-trained PPO model is then selected, with fallback options implemented when direct parameter matches are unavailable.

\textbf{Episode Simulation.} For each parameter set, we simulate multiple episodes (typically 100) using unique random seeds to ensure trajectory diversity. Within each episode, states are systematically recorded and input to the PPO networks, actions are computed and executed, and rewards are tracked until episode termination at time $T$.

\textbf{Data Aggregation and Storage.} The collected trajectories encompass states, actions, rewards, episode termination indicators, and comprehensive metadata including implementation shortfall, final inventory positions, and scenario labels. The data is efficiently stored using Zarr archives with Blosc compression to optimize storage requirements and access speed.

This systematic approach to PPO-based trajectory generation significantly enhances the FlowOE training corpus, thereby promoting robust imitation learning performance across diverse market scenarios and parameter regimes.

\subsection{State-Dependent Trading Strategy}

Due to the complexity of solving the HJB equation analytically, we implement a state-dependent trading strategy that captures the key features of the optimal solution:

\begin{equation}
\nu_t^* = \Phi(t, X_t, S_t, V_t) \cdot \Psi(V_t, \mathbb{E}[V_T|V_t]) \cdot \Omega(T-t)
\end{equation}

where:
$\Phi(t, x, s, v) = \left(\frac{x}{T-t}\right)^{\frac{1}{1+\beta}}$ is the base trading rate accounting for non-linear impact
$\Psi(v, \mathbb{E}[v_T]) = \left(\frac{\mathbb{E}[V_T|V_t]}{V_t}\right)^{0.5}$ is the volatility adjustment factor
$\Omega(\tau) = \min(2, \frac{1}{\tau})$ is the urgency factor as time to maturity decreases

The expected future variance is computed using the mean-reverting property:

\begin{equation}
\mathbb{E}[V_T | V_t] = \theta + (V_t - \theta)e^{-\kappa(T-t)}
\end{equation}

We define the problem mathematically and present four expert strategies: TWAP, VWAP, an approximated Almgren-Chriss strategy (AC-Approx), and a Heston-Optimal strategy tailored to the non-linear impact and stochastic volatility setting. The mathematical formulations are derived, and their performance is assessed through simulations.

\subsection{Discretization}
In discrete time, for step $k = 0, 1, \dots, N-1$, the trader sells $x_k = \nu_k \Delta t$ shares. The price and variance are updated using a Milstein scheme for the variance to ensure non-negativity:
\begin{align}
    \log S(t_{k+1}) &= \log S(t_k) + \left( \mu - \frac{1}{2} V(t_k) - \eta \nu_k \right) \Delta t + \sqrt{V(t_k)} \sqrt{\Delta t} Z_{S,k}, \\
    V(t_{k+1}) &= V(t_k) + \kappa \left( \theta - V(t_k) \right) \Delta t + \xi \sqrt{V(t_k)} \sqrt{\Delta t} Z_{v,k} + \frac{\xi^2}{4} \Delta t \left( Z_{v,k}^2 - 1 \right),
\end{align}
where $Z_{S,k}$ and $Z_{v,k}$ are correlated standard normal random variables with correlation $\rho$. The execution price is:
\begin{equation}
    \tilde{S}_k = S(t_k) - \epsilon \nu_k^\beta.
\end{equation}
Inventory and cash updates are:
\begin{align}
    q(t_{k+1}) &= q(t_k) - x_k, \\
    C(t_{k+1}) &= C(t_k) + x_k \tilde{S}_k.
\end{align}

\section{MDP Formulation of Optimal Execution}

We formulate optimal execution as an MDP with tuple $(\mathcal{S}, \mathcal{A}, \mathcal{R}, \mathcal{P}, \gamma)$.

\textbf{State Space $\mathcal{S}$:} At time $t$, the state is $s_t = (\tau_t, q_t, S_t, V_t)$ where:
\begin{itemize}
\item $\tau_t = T - t$: remaining time (normalized by total time $T$)
\item $q_t$: current inventory (normalized by initial inventory $X_0$)
\item $S_t$: mid-price (normalized by initial price $S_0$)
\item $V_t$: variance (normalized by initial variance $V_0$)
\end{itemize}

\textbf{Action Space $\mathcal{A}$:} The action $a_t \in [0, 1]$ represents the fraction of remaining inventory to trade at time $t$.

\textbf{Reward ($\mathcal{R}$):} The agent aims to maximize cumulative rewards. The immediate reward $r_k$ at step $k$ can be defined as the cash received from selling $n_k$ shares, penalized by a risk term. A common objective in Optimal Execution (OE) is to minimize the Implementation Shortfall (IS) and its variance. We define the reward to align with this objective:
    \begin{equation}
        r_k(s_k, a_k, s_{k+1}) = n_k \tilde{S}_k - \lambda \cdot \text{RiskTerm}_k
        \label{eq:reward_function}
    \end{equation}
    where:
    \begin{itemize}
        \item $n_k$ is the number of shares sold in step $k$.
        \item $\tilde{S}_k$ is the actual execution price per share at step $k$.
        \item $\lambda$ is a risk aversion coefficient.
        \item $\text{RiskTerm}_k$ is a term quantifying the risk incurred during step $k$, which could be related to price volatility or deviation from a target execution schedule.
    \end{itemize}
    A significant negative reward (penalty) is applied at the terminal step if any shares $q_N$ remain unliquidated ($q_N > 0$). The overall objective for the agent is to maximize the sum of discounted rewards $\sum_{k=0}^{N-1} \gamma^k r_k$. This reward structure encourages the agent to maximize execution proceeds while managing risk, effectively aiming to minimize an objective similar to the Almgren-Chriss cost function, i.e., $\mathbb{E}[\text{IS}] + \lambda \text{Var}[\text{IS}]$.
where $\text{trading\_cost}_t = \varepsilon |\nu_t|^\beta \cdot \text{shares\_traded}_t$ and $\lambda$ is the terminal inventory penalty.

\textbf{Transition Dynamics $\mathcal{P}$:} The market follows Heston stochastic volatility dynamics:
\begin{align}
dS_t &= \mu S_t dt + \sqrt{V_t} S_t dW_t^S - \eta \nu_t S_t dt \\
dV_t &= \kappa(\theta - V_t)dt + \xi\sqrt{V_t} dW_t^V
\end{align}
where $\nu_t$ is the trading rate, $\eta$ is the permanent impact coefficient, and $\langle dW_t^S, dW_t^V \rangle = \rho dt$.

The inventory evolves as: $q_{t+1} = q_t - a_t \cdot q_t$

\textbf{Discount Factor:} $\gamma = 0.99$

\subsection{Gymnasium Environment Implementation}

We implement the execution environment following OpenAI Gymnasium standards:

\begin{algorithm}[H]
\caption{MDP Approximation of \texttt{OptimalExecutionHestonEnv}}
\label{alg:heston-mdp}
\begin{algorithmic}[1]
\REQUIRE Environment parameters $(T,N,X_0,S_0,V_0,\mu,\kappa,\theta,\xi,\rho,\eta,\epsilon,\beta,\lambda,\phi)$ where $\lambda$: risk-aversion; $\phi$: terminal-inventory penalty
\STATE $\Delta t \gets T/N$

\STATE \textbf{Function Reset():}
\STATE \hspace{1em} $t_0\gets 0$, $q_0\gets X_0$, $S_0\gets S_0$, $V_0\gets V_0$, $C_0\gets 0$
\STATE \hspace{1em} \textbf{return} $\mathbf{s}_0 = f_{\text{norm}}(t_0,q_0,S_0,V_0)$

\STATE \textbf{Function Step}($a_t \in [0,1]$): \COMMENT{$a_t$ = fraction of remaining inventory to trade}
\IF{$t = T-\Delta t$} 
\STATE $a_t \gets 1$ \COMMENT{force full liquidation in the last step}
\ENDIF
\STATE $\text{shares} \gets a_t \cdot q_t$
\STATE $\nu_t \gets \text{shares}/\Delta t$ \COMMENT{trading rate}
\STATE $\text{temp} \gets \epsilon \cdot \nu_t^{\beta}$ \COMMENT{non-linear temporary impact}
\STATE $\text{perm} \gets \eta \cdot \nu_t$ \COMMENT{permanent impact per unit time}
\STATE $P_t^{\text{exec}} \gets S_t-\text{temp}-0.5\eta \cdot \text{shares}$
\STATE $\text{cost}_t \gets \text{temp} \cdot \text{shares}+0.5\eta \cdot \text{shares}^2$
\STATE $C_{t+\Delta t} \gets C_t + P_t^{\text{exec}} \times \text{shares}$
\STATE $q_{t+\Delta t} \gets q_t-\text{shares}$ \COMMENT{update inventory}

\STATE Draw $Z_1,Z_2\sim\mathcal{N}(0,1)$
\STATE $W_S\gets Z_1$, $W_V\gets \rho Z_1+\sqrt{1-\rho^2} \cdot Z_2$ \COMMENT{correlated Brownian motions}
\STATE $dV \gets \kappa(\theta-V_t)\Delta t + \xi\sqrt{V_t}\sqrt{\Delta t} \cdot W_V + \frac{1}{4}\xi^2\Delta t \cdot (W_V^2-1)$
\STATE $V_{t+\Delta t}\gets\max(0,V_t+dV)$
\STATE $d\ln S\gets(\mu-\frac{1}{2} V_t)\Delta t+\sqrt{V_t}\sqrt{\Delta t} \cdot W_S$
\STATE $S_{t+\Delta t}\gets S_t \cdot e^{d\ln S}-\text{perm}$

\STATE $t \gets t+\Delta t$
\STATE $\text{done}\gets(t\ge T) \lor (q_{t+\Delta t}\le10^{-6})$
\STATE $\text{hold}\gets-\lambda \cdot q_{t+\Delta t}^{2} \cdot V_{t+\Delta t} \cdot \Delta t$
\STATE $\text{reward}\gets -\frac{\text{cost}_t}{X_0S_0}\times100 + \text{hold}$
\IF{$\text{done} \land q_{t+\Delta t}>0$} 
\STATE $\text{reward}\gets\text{reward}-\phi \cdot (q_{t+\Delta t}/X_0)$ \COMMENT{penalise leftover inventory}
\ENDIF
\STATE \textbf{return} $f_{\text{norm}}(t,q_{t+\Delta t},S_{t+\Delta t},V_{t+\Delta t})$, $\text{reward}$, $\text{done}$, $\text{info}$

\end{algorithmic}
\end{algorithm}

\begin{table}[H]
    \centering
    \caption{MDP Components of \texttt{OptimalExecutionHestonEnv}}
    \label{tab:heston-mdp}
    \renewcommand{\arraystretch}{1.5}
    \small
    \begin{tabular}{@{}p{2.8cm}p{7.5cm}p{4.2cm}@{}}
        \toprule
        \textbf{Element} & \textbf{Definition} & \textbf{Source-code reference} \\ 
        \midrule
        
        \textbf{State} $\mathbf{s}_t$ &
        Normalised Tuple: $\bigl(\text{time left},\;\text{inventory ratio},\;\log S_t,\;\log\sqrt{V_t}\bigr)$
        & \texttt{get\_obs()} \\
        \addlinespace[4pt]
        
        \textbf{Action} $a_t$ &
        Continuous in $[0,1]$; trades $a_t \times q_t$ shares at each step
        & \texttt{step()} \\
        \addlinespace[4pt]
        
        \textbf{Dynamics} &
        \begin{minipage}[t]{7.3cm}
        \vspace{1pt}
        $\bullet$ Heston SV model for asset price\\
        $\bullet$ CIR--Milstein scheme for variance\\
        $\bullet$ Linear permanent \& nonlinear temporary market impact
        \vspace{1pt}
        \end{minipage}
        & \texttt{step()} lines on $dV$, $d\ln S$, impacts \\
        \addlinespace[4pt]
        
        \textbf{Reward} &
        $-\,$(execution cost  $+\,$ inventory-risk penalty $+\,$ terminal leftover penalty)
        & reward computation \\
        \addlinespace[4pt]
        
        \textbf{Termination} &
        End of horizon $t=T$ \textbf{or} inventory $\approx 0$
        & \texttt{done} flag \\ 
        
        \bottomrule
    \end{tabular}
\end{table}

\textbf{Key Environment Features:}
\begin{itemize}
\item Time discretization: 100 steps over $T = 1.0$
\item Non-linear temporary impact: $f(\nu) = \varepsilon |\nu|^\beta$ with $\beta \in [0.3,0.5, 0.8]$
\item Linear permanent impact proportional to trading rate
\item Terminal constraint: complete liquidation required ($q_T = 0$)
\end{itemize}

\newpage

\end{appendix}

\bibliographystyle{elsarticle-harv}

\bibliography{references}

\end{document}